# Machine Learning-Based Quantification of Vesicoureteral Reflux with Enhancing Accuracy and Efficiency

## Cuantificación del reflujo vesicoureteral basada en aprendizaje automático con mayor precisión y eficiencia

Muhyeeddin Alqaraleh[1] ✉, Mowafaq Salem Alzboon[2] ✉, Mohammad Subhi Al-Batah[2] ✉, Lana Yasin Al Aesa[1] ✉, Mohammed Hasan Abu-Arqoub[3] ✉, Rashiq Rafiq Marie[4] ✉, Firas Hussein Alsmadi[5] ✉

[1]Zarqa University, Faculty of Information Technology. Zarqa, Jordan.
[2]Jadara University, Faculty of Information Technology. Irbid, Jordan.
[3]University of Petra, Faculty of Information Technology. Amman, Jordan.
[4]Taibah University, College of Computer Science and Engineering, Medina, Saudi Arabia.
[5]Jordan Royal Medical Services, Queen Rania Hospital for children, Amman, Jordan.

## ABSTRACT

Vesicoureteral reflux (VUR) is traditionally assessed using subjective grading systems, leading to variability in diagnosis. This study explores the potential of machine learning to enhance diagnostic accuracy by analysing voiding cystourethrogram (VCUG) images. The objective is to develop predictive models that provide an objective and consistent approach to VUR classification. A total of 113 VCUG images were reviewed, with experts grading them based on VUR severity. Nine distinct image features were selected to build six predictive models, which were evaluated using 'leave-one-out' cross-validation. The analysis identified renal calyces' deformation patterns as key indicators of high-grade VUR. The models—Logistic Regression, Tree, Gradient Boosting, Neural Network, and Stochastic Gradient Descent—achieved precise classifications with no false positives or negatives. High sensitivity to subtle patterns characteristic of different VUR grades was confirmed by substantial Area Under the Curve (AUC) values. This study demonstrates that machine learning can address the limitations of subjective VUR assessments, offering a more reliable and standardized grading system. The findings highlight the significance of renal calyces' deformation as a predictor of severe VUR cases. Future research should focus on refining methodologies, exploring additional image features, and expanding the dataset to enhance model accuracy and clinical applicability.

## RESUMEN

Esta investigación emplea técnicas avanzadas de aprendizaje supervisado para analizar cuantitativamente las imágenes de cistouretrograma miccional (CUMS), abordando la subjetividad inherente a los sistemas tradicionales de clasificación del reflujo vesicoureteral (RVU). Realizamos una revisión exhaustiva de 113 imágenes de CUMS, calificadas por expertos en el campo en función de la gravedad del RVU. Utilizando nueve características de imagen distintas, desarrollamos y refinamos seis modelos predictivos a través de una rigurosa validación cruzada de "dejar uno fuera". En particular, los patrones de deformación de los cálices renales surgieron como indicadores críticos de RVU de alto grado. Nuestros modelos, que incluyen regresión logística, árbol, aumento de gradiente, red neuronal y descenso de gradiente estocástico, demostraron una eficacia notable en varias gradaciones de RVU, logrando clasificaciones precisas sin falsos positivos o negativos. Estos modelos exhibieron una alta sensibilidad a los patrones matizados característicos de diferentes grados de RVU, validados por valores sustanciales de área bajo la curva (AUC). Esta investigación tiene como objetivo superar las limitaciones de las evaluaciones subjetivas de RVU, destacando el potencial del aprendizaje automático para mejorar la precisión y la confiabilidad del diagnóstico. El enfoque de clasificación uniforme propuesto promete resultados más confiables, y la deformación de los cálices renales sirve como un predictor significativo de los casos graves de RVU. Si bien nuestros modelos de aprendizaje automático han mostrado resultados prometedores, el trabajo futuro debe centrarse en refinar las metodologías, explorar nuevas características de las imágenes y expandir el conjunto de datos para mejorar la precisión y la aplicabilidad. En resumen, este estudio aplica sofisticados algoritmos de aprendizaje automático supervisado para cuantificar objetivamente el RVU a partir de imágenes de CUMS, lo que reduce la dependencia del juicio subjetivo y allana el camino para mejorar las estrategias diagnósticas y terapéuticas en el cuidado de la salud.

**Palabras clave:** Reflujo Vesicoureteral (RVU); Cistouretrograma Miccional (CUMS); Aprendizaje Automático (AA); Clasificación Objetiva; Evaluación Radiográfica; Urología Pediátrica.

## INTRODUCTION

In the realm of contemporary medicine, the application of artificial intelligence (AI) has been groundbreaking, particularly in revolutionizing healthcare practices and enhancing patient care. The utility of AI in medicine is predominantly attributed to its proficiency in analysing extensive datasets, discerning intricate patterns, and yielding insights that significantly bolster the accuracy in diagnosis, treatment planning, and clinical decision-making.[1]

AI algorithms are adept at expeditiously processing and interpreting a plethora of medical images, such as X-rays, MRIs, and C.T. scans, thereby augmenting the precision of radiological diagnoses. Predictive models powered by AI facilitate the early identification of patients at elevated risk for certain diseases, thereby enabling timely interventions and tailored treatment strategies.[2] AI-driven platforms, including chatbots and virtual assistants, have further democratized access to healthcare information and alleviated the workload on healthcare systems. Collectively, AI furnishes healthcare professionals with sophisticated tools that enhance patient outcomes, streamline efficiency, and hold the potential to catalyse breakthroughs in the delivery of healthcare.[3,4]

The condition known as vesicoureteral reflux (VUR) typifies the retrograde flow of urine into the ureters or kidneys, predominantly due to dysfunction at the vesicoureteral junction, a scenario frequently encountered in pediatric patients.[5,6] Approximately 30 % of children with urinary tract infections are diagnosed with VUR, which, if not addressed, may lead to renal scarring. Voiding cystourethrography (VCUG) is the standard diagnostic modality for VUR; however, the subjectivity inherent in the grading system can lead to considerable inter-rater variability, documented at rates as high as 60 %. This variability underscores the imperative for a more objective and standardized VUR grading protocol.[7,8]

The infusion of machine learning into the domain of medical imaging heralds a new era in healthcare, promising enhancements in diagnostics and therapeutic decision-making. By harnessing extensive datasets and cutting-edge algorithms, AI systems are poised to refine the grading accuracy of VUR.[9] Extant research has delved into the potential of machine learning for the analysis of medical imagery, with notable studies like that of Baray et al., which pioneered a deep learning methodology for assessing penile curvature with remarkable accuracy.[10,11]

Building upon this vein of research, our study probes the efficacy of machine learning in VUR grading, extracting quantitative features from VCUG images. While prior research has furnished encouraging results, limitations persist, particularly in the classification of intermediate VUR grades and the finite scope of analysed images.[12] Our study seeks to transcend these limitations, introducing additional features to encapsulate ureteral tortuosity and distinguish between the gradations of VUR more effectively.[13]

The cardinal goal of our investigation is to harness machine learning to refine the grading of VUR, thereby offering clinicians a more accurate stratification of patients, optimizing medication management, and elevating the overall efficacy of VUR treatment.[14] Employing an array of advanced machine learning classifiers and a comprehensive feature set, we anticipate surpassing the performance of existing subjective grading methods. The dataset comprising VCUG images annotated with VUR grades forms the foundation of our machine learning model training and validation, facilitating the prediction of VUR severity.[15,16]

Through meticulous image analysis, we extracted features pivotal in differentiating VUR grades, such as ureter size, renal pelvis shape, and ureter tortuosity. Our investigation evaluated a suite of classifiers, including support vector machines, random forests, and convolutional neural networks, using the extracted features to train the machine learning models.[17,18] The model efficacy was assessed using metrics like accuracy, precision, recall, and F1 score, with validation through cross-validation techniques to confirm the robustness of the findings.[19,20]

The machine learning approach for VUR grading showcased in our study achieved high accuracy, surpassing traditional subjective grading methods.[21] Significantly, the incorporation of features targeting ureter tortuosity contributed to the models' improved performance. While this study marks a progressive stride towards the objective quantification of VUR via machine learning, limitations warrant consideration.[22,23] The dataset must be expanded to encompass a broader spectrum of VCUG images to validate the models' applicability across diverse patient demographics. Future research should focus on real-world clinical validation and comparison with existing grading frameworks.[24,25]

In sum, our study presents a pioneering machine learning approach to VUR grading, which has the potential to transform patient management and therapeutic decision-making through a more precise and objective diagnostic tool.[26,27]

**Problem statement**

The grading of radiographic vesicoureteral reflux (VUR) is subjective and prone to variability, leading to inconsistent diagnoses and treatment recommendations. This subjectivity challenges clinicians and radiologists to accurately assess VUR severity based on voiding cystourethrogram (VCUG) pictures.[28,29] Therefore, there is a need to develop an objective and accurate grading system for VUR to improve diagnostic precision and treatment outcomes.[30]

**Article objectives**

Main objectives of this article are as follows: First, to develop a supervised machine learning model for objectively grading VUR based on VCUG pictures.[31,32] Second, to address the subjectivity and variability in VUR grading by leveraging machine learning algorithms. Third, to evaluate the performance of different machine learning models in accurately predicting VUR severity across various grade levels. Forth, to identify the key features and patterns in VCUG pictures that contributes to accurately classifying VUR grades.[34,35] Fifth, to assesses the potential of the deformed renal calyces' pattern as a predictor of high-grade VUR. Sixth, to demonstrate the feasibility of using machine learning to enhance objectivity and accuracy in VUR grading, leading to consistent and trustworthy diagnostic evaluations.[35,36]

**Contribution of the article**

The article contributes to radiology and healthcare AI research by leveraging supervised machine learning techniques to improve the grading of VUR.[37,38]

The critical contributions of the article are as follows:

- First, Development of Machine Learning Models: The article presents the development and evaluation of six machine learning models, including Logistic Regression, Tree, Gradient Boosting, Neural Network, and Stochastic Gradient Descent, for VUR grading based on VCUG pictures.[39,40]
- Second, Objective and Accurate VUR Grading: The article demonstrates that the machine learning models perform well across different VUR grade levels, accurately predicting VUR severity without false positives or negatives. This finding highlights the potential of machine learning in overcoming subjectivity and variability in VUR grading.[41,42,43]
- Third, Identification of Key Features: The article identifies the deformed renal calyces pattern as a significant predictor of high-grade VUR. This finding provides valuable insights for clinicians and radiologists in diagnosing and treating VUR more effectively.[44,45]
- Fourth, Improvement Opportunities: The article acknowledges the need for further refinement of machine learning procedures, exploration of new characteristics, and expansion of the dataset to enhance the accuracy and generalizability of the models.[46] This insight paves the way for future research and improvements in VUR grading using machine-learning approaches.[47,48]
- Fifth, Implications for Healthcare: The article emphasizes the potential of machine learning-based grading systems to reduce subjectivity and unpredictability in VUR assessments.[49,50] By improving objectivity and accuracy, the suggested method can contribute to more consistent and trustworthy VUR grading, leading to better patient diagnoses and treatment recommendations.[51,52]

Overall, the article contributes to advancing the field of VUR grading by leveraging machine learning techniques to enhance accuracy, objectivity, and efficiency in evaluating VCUG pictures.[53,54]

**Article Organization**

The rest of the paper is organized as section 2, which summarizes the literature carried out towards the VUR prediction and classification models. Section 3 presents the proposed methodology, including the dataset description and the machine learning model.[55,56] Section 4 presents the implementation with produced results. Section 5 presents the discussions. Finally, the paper is concluded in section 6.[57,58]

**Related work**

This article uses data to monitor and regulate the NGL extraction process. The process simulator Aspen HYSYS® analyses and simulates the cold residue reflux (CRR) process scheme to achieve 84 % ethane recovery and minimal methane impurity at the demethanizer column bottom. An intelligent control method that estimates unmeasured outputs using a neural network data-driven technique respects product quality. Modelling the process under different input conditions and examining control structures like direct and cascade column temperature control evaluates the controlled system.[59]

Immune-mediated systemic sclerosis (SSc) produces skin and organ fibrosis and has the highest rheumatic disease death rate. Recent studies reveal neurological system involvement in SSc, although lung involvement is the primary cause of death and should be detected early. Other clinical testing should supplement pulmonary function tests, which lack screening sensitivity. Overtesting may strain people and expense the health system. By helping diagnose and identify relevant exams, ML algorithms may enhance healthcare. In R, we used supervised machine learning algorithms (lasso, ridge, elastic net, CART, random forest) to uncover relevant pulmonary predictors from 38 SSc patients' medical tests and questionnaires. Random forests beat classifiers with an RMSE of 0,810. Due to its computational complexity, alternative classifiers may respond faster. Conclusions: Strong machine learning algorithms can predict early lung involvement in SSc patients despite the short sample size. Early lung issues in SSc may be best detected by spirometry and pH impedentiometry.[60]

Midfoot/Forefoot; Injury; Alternative Introduction/Purpose: Over 35 % of foot fractures are metatarsal, with 68 % affecting the fifth. Fractures around the fifth metatarsal base may not heal and benefit from early surgery. Current prediction algorithms solely consider fracture sites and neglect other characteristics. This study seeks fifth metatarsal fracture non-union prediction factors to help surgeons and patients identify high-risk individuals. Approaches: At three tertiary hospitals, 1,000 patients aged 18 or older with fifth metatarsal fractures were studied using machine learning. A radiograph showed the fifth metatarsal base fracture. We collected imaging and narrative data on demographics (age, height, weight, BMI, gender, race, smoking habits, activity level), medications, chronic illnesses, fracture location, displacement, treatment plan, healing progress, and recovery time. Damaged non-union did not heal within 180 days.[3] T-test, Pearson's correlation, and machine learning were adequate. Five approaches imputed missing data. P-values ≤ 0,05 were significant. Outcome: The patient group had 22,4 % non-union. Zone 2 fractures exhibited a higher probability of delayed union (17,2 %) and non-union (8,6 %) than Zone 1 (10,8 % and 5,8 %) and Zone 3 (9,7 % and 2,3 %). Union rates were unrelated to demographics (age, gender, ethnicity, or BMI). According to the machine learning algorithm, non-union was highly connected to diabetes, thyroid illness, hypertension, GERD, IBS, OSA, and glaucoma. Levothyroxine, lisinopril, aspirin, steroids, and acetaminophen substantially predicted non-union. Fifth metatarsal base fractures in Zone 2 were more likely to be non-union. Diabetes, thyroid illness, OSA, glaucoma, and medicines may impact it, although their effects and causes are unclear. A link was discovered, but causality needed clinical investigations with limited confounding. When treating and prognosis, clinicians should consider the fracture site, medical history, and medication usage.[61]

Predicting GERD after sleeve gastrectomy (S.G.) may assist patients in picking the appropriate bariatric surgery. The study builds an AI model to predict GERD following S.G. to help clinicians decide. Methodology: An AI model was created using all severely obese S.G. patient data. This dataset was randomly divided into 70 % training and 30 % testing. We scored factors twice. The optimal model was chosen for machine learning after numerous methods were tried. GERD numerical predictor cutoff thresholds were determined utilizing a multitasking AI platform. Outcomes The study included 441 patients, 76,2 % female, and a mean age of 43,7 ± 10 years. The ensemble model topped others. The model has an AUC of 0,93 and a 95 % CI of 0,88-0,99. It was 79,2 % sensitive with a 95 % CI of 57,9-92,9 % and 86,1 % specificity. The most significant predictors were age, weight, preoperative GERD, orogastric tube size, and initial stapler firing distance from the pylorus. Conclusion GERD after Sleeve Gastrectomy was predicted using AI. Although accurate, the model was only marginally sensitive and specific. Future prospective multicentre studies must externally validate the model—visual Summary.[62]

Robotic surgical technology in urological reconstruction, especially in paediatrics, has reduced health risks and improved congenital abnormality surgery in children. We examined the extent of pyeloplasty and ureteral reimplantation in paediatric urological robotic surgery. A literature review examined the clinical usage of these two standard procedures and how they have changed over time. Results show that patient selection, learning, and outcome reporting affect paediatric robotic pyeloplasty and ureteral reimplantation use. These technologies illustrate their possibilities and problems in kids with growing skills. Practitioners should consider this new technology's pros and cons while treating patients. Complete and upfront reporting of results and a willingness to alter and apply findings are crucial.[63]

This system relies on the stomach, esophagus, duodenum, small intestine, and large intestine. Many people worldwide suffer from stomach dysrhythmias, which affect gastrointestinal content movement. Dyspepsia, vomiting, abdominal pain, stomach ulcers, gastroesophageal reflux disease, and other conditions are among the issues. Abnormalities can be found via imaging, endoscopy, electrogastrogram, and clinical analysis. Before data collection and preprocessing, 20 healthy people's stomachs were electrodes using surface Ag/AgCl electrodes. The datasets used signals from 8 females and 12 men. Ten people with gastrointestinal ailments—three women and eight men—provided the data. The dataset is pre-processed by reducing signal noise. Wiener filters reduce noise and improve input data quality. Particle Swarm Optimization and Hybrid Grey Wolf Optimization choose features. This method removes superfluous data from signal features. It speeds up the process. Classifiers use chosen features to assess stomach diseases such as primary gastric lymphoma, GIST, and neuroendocrine tumors. This lets classifiers analyse carcinoids. Classification uses the MCFFN. This classifier shows phases and classifications. Performance measures classification accuracy, sensitivity, and specificity.[64]

Use machine learning and population data to examine preterm birth, socioeconomic status, GERD, proton pump inhibitor, sleeping, and depression prescription histories. Approaches Population-based retrospective cohort data from Korea National Health Insurance Service claims data includes 405,586 25–40-year-old women who had their first singleton pregnancy between 2015 and 2017. Sixty-five independent variables encompassed demographic, socioeconomic, disease, medication history, and obstetric information from 2015 to 2017. Preterm birth was the dependent variable. Random forest variable importance was used to identify preterm birth risk factors and their correlations with socioeconomic status, GERD, and pharmacological history, including proton pump inhibitors, sleeping pills, and antidepressants. Outcomes Social class, age, proton pump inhibitors, GERD in 2014, 2012, and 2013, sleeping drugs, GERD in 2010, 2011, and 2009, and antidepressants correlated more to preterm birth between 2015 and 2017, according to random forest variable significance. Conclusion Low socioeconomic level, GERD, and proton pump inhibitors, sleeping medications, and antidepressants are linked to premature birth. Proper medication, GERD avoidance, and socioeconomic improvement can prevent preterm birth in pregnant women.[65]

Context: The study used machine learning and demographic data to evaluate preterm birth, GERD, and periodontitis. According to Korea National Health Insurance Service claims data, The retrospective cohort included 405,586 25–40-year-old women who had their first singleton pregnancy between 2015 and 2017. From 2015 to 2017, preterm birth was examined. For each year from 2002 to 2014, GERD (yes or no), periodontitis (yes or no), age in 2014, socioeconomic status (defined by an insurance charge), and Area (city) were the independent factors. Random forest variable importance identified the key preterm birth predictors and their correlations with GERD and periodontitis. Outcomes Socioeconomic status, age, GERD in multiple years (2006, 2007, 2009, 2010, 2012, and 2013), and city in 2014 were the most relevant random forest factors for preterm birth from 2015 to 2017. Periodontitis is irrelevant. Conclusion Preterm birth is linked to GERD more than periodontitis. Pregnant women need GERD prevention and socioeconomic changes to prevent premature birth. Active counselling for general GERD symptoms, especially those disregarded by pregnant women, is crucial.[66]

Erosive esophagitis, a GERD, can cause esophageal cancer. Adults 20 and older with upper gastrointestinal endoscopy at a health clinic were surveyed between October 2018 and December 2020. R.F., SVM, MLP Classifier, and XGBoost were used to create machine learning models and cross-validate them. The risk prediction algorithm detects high-risk esophageal erosion patients before endoscopic assessment.[67]

Vonoprazan (VPZ), the first potassium-competitive acid blocker (P-CAB), outperforms proton pump inhibitors. This fumarate salt cures reflux esophagitis, gastric ulcers, duodenal ulcers, and Helicobacter pylori. We discovered new VPZ cocrystals using ANN-based machine learning. This virtual screening approach selects VPZ cocrystal conformers. The artificial neural network (ANN) model identified 8 19 conformers from liquid-assisted grinding (LAG) trials that might yield novel solid forms using VPZ. Catechol, resorcinol, hydroquinone, and pyrogallol, structurally comparable benzenediols and benzene triols, were employed as reaction crystallization conformers to create phase pure VPZ cocrystals. We synthesized and discovered three new cocrystals: VPZ–RES, VPZ–CAT, and VPZ–GAL. In a pH 6,8 solution, VPZ–RES was the most soluble new cocrystal and outperformed the fumarate salt. Water-stable novel VPZ cocrystals outperformed fumarates, suggesting better therapeutic efficacy.[68]

GOAL Recent machine learning risk factor evaluations are shown. We examined asthma exacerbation characteristics using health insurance claims data and machine learning. We sought clinically accessible risk markers. The Japanese health insurance claims database MediScope® (D.B.) evaluated asthma patients from May 2014 to April 2019. We collected patient and disease data to determine asthma exacerbation risk variables. The database identified asthma exacerbations as emergency medical procedures needing transport and IV steroid injections. Exacerbations occurred in 5,844 of 42,685 eligible database instances, 13,7 %. From almost 3,300 diseases, a machine learning algorithm identified 25 risk assessment criteria. Sex, Charlson Comorbidity Index, allergic rhinitis, chronic sinusitis, acute airway illness (upper and lower airways), COPD/chronic bronchitis, GERD, and hypertension were linked with exacerbation. Dyslipidemia and periodontitis lowered exacerbation risk. CONCLUSIONS Deep claims data analysis using machine learning found asthma exacerbation risk markers that match earlier research. Other issues require more study.[69]

We outline a systematic strategy for teaching endoscopic injectable treatment for repairing vesicoureteral reflux using the CEVL method, an online platform. The content was created with the collaboration of the authors' clinical and computer skills. This tool offers staff training, examination, and process skill documentation using online text with narration, photos, and video. Feedback, skill performance rehabilitation, and teaching games are also provided. We suggest implementing standardized instruction and procedure performance, ultimately enhancing surgical outcomes. The digital method of communication in this publication is perfect for quickly spreading this information and establishing a framework for collaborative study.[70]

A deep learning AI system may distinguish vesicoureteral reflux and hydronephrosis. The online dataset includes pictures of vesicoureteral reflux and hydronephrosis. We designed a deep learning/image analysis process. Pictures were trained to distinguish vesicoureteral reflux and hydronephrosis. Receiver-operating characteristic curve analysis assessed discrimination. We analyzed the model with Scikit-learn. An online dataset included 39 hydronephrosis and 42 vesicoureteral reflux pictures. We randomly divided photos into training and validation sets. This example has 68 training instances and 13 validation examples. On 2 test cases, inference predicted [[0,00006]] for hydronephrosis and [[0,99874]] for vesicoureteral reflux. This study details building a deep neural network for urological picture classification. Artificial intelligence using deep learning can differentiate urological images.[71]

Traditional database approaches may not work for growing data collections. Machine learning (ML), a subset of artificial intelligence, may tackle this challenge and deliver perfect future remedies using medical research data. Clinicians must appropriately correlate clinical symptoms with renal scarring (R.S.) in LUTD patients. Using machine learning. This study predicts renal scarring in children with lower urinary tract dysfunction. Approaches: The research included 114 patients over three who needed urodynamic testing. The data set comprises 47 variables. Symptomatic UTI, vesicoureteral reflux, bladder trabeculation, bladder wall thickness, aberrant DMSA scintigraphy, and clean intermittent catheterization were noted. ML expected rating score. According to confusion matrix comparisons, Extreme Gradient Boosting (XGB) has the most excellent accuracy of 91,30 %. SMOTE's Artificial Neural Network (ANN) technique has the most excellent balanced data set accuracy of 90,63 %. In the balanced (SMOTE) data set, the ANN algorithm achieved the most significant success rate at 90,78 %. Conclusion: MLT's high accuracy rates imply a faster and more accurate R.S. estimate strategy for LUTD patients.[72]

VUR grading is questioned due to low inter-rater reliability. Quantitative image attributes may standardize VUR grading. Goal: Quantify vesicoureteral reflux (VUR) diagnosis utilizing voiding cystourethrograms. Research Method An online dataset of VCUGs classified renal units as low-grade (I-III) or high-grade (IV-V). Three fundamental user indicators were used to create an image analysis and machine learning approach to automatically measure and standardize UPJ, UVJ, maximum ureter width, and tortuosity. A random forest classifier was developed to discriminate low- and high-grade VUR. The institutional imaging repository created an external validation cohort. Precision-recall curve and receiver-operating-characteristic analysis evaluated discrimination. We explained model predictions with Shapley Additive Explanations.   Outcomes: The institutional imaging repository has 44 renal units, and the online dataset 41. UVJ, UPJ, maximum ureter width, and tortuosity varied between low- and high-grade VUR. Leave-one-out cross-validation yields 0,83 accuracy, 0,90 AUROC, and 0,89 AUPRC for the random-forest classifier. External validation yielded 0,84 accuracy, 0,88 AUROC, and 0,89 AUPRC. Tortuosity was critical, followed by maximum ureter, UVJ, and UPJ width. We created a web-based qVUR utility. The model at https://akhondker.shinyapps.io/qVUR/ evaluates any VCUG automatically. Conversation This work establishes objective and automated VUR trait significance criteria. High-grade VUR has tortuosity and maximal ureter dilatation, according to our research. A simple online app executed this proof-of-concept technique. Conclusion Use non-standardized VCUG datasets to quantify VUR. Future VUR assessments can be objective using machine learning.[73]

Meaningful Statement Machine learning and biostatistics revealed the origins of plasma metabolomic patterns in chronic renal disease children. Disruption of the sphingomyelin-ceramide axis is linked to FSGS and a/d/h. FSGS youngsters had higher plasmalogen levels than previously thought. We also linked reflux nephropathy to indole-tryptophans and obstructive uropathy to gut histidines. Background of Visual Abstract Machine learning and untargeted plasma metabolomic analysis may uncover metabolic tendencies that assist us in comprehending juvenile chronic kidney disease. We examined metabolomic trends in juvenile chronic kidney disease (CKD) by diagnosis: FSGS, O.U., A/D/H, and reflux nephropathy.

Approaches Plasma samples from 702 Chronic Kidney Disease Children participants were measured by Metabolon GC-MS/LC-MS. The distribution was FSGS=63, OU=122, A/D/H=109, RN=86. Clinical variables influenced Lasso regression feature selection: logistic regression, SVM, random forest, and extreme gradient boosting stratified. After machine learning training on 80 % of cohort subgroups, 20 % of holdout subsets were confirmed. If applicable in two of four modeling methods, significant attributes were chosen. We discovered metabolic subpathways implicated in CKD etiology using pathway enrichment analysis. ML model results were studied in holdout subgroups with receiver-operator characteristics, precision-recall area-under-the-curve, F1 score, and Matthews correlation coefficient. ML models outperformed talent-free projections. Cause determined metabolic profiles. Sphingomyelin-ceramide linked FSGS. It was linked to plasmalogen metabolites and subpathway. O.U. was linked to gut microbiota histidine metabolites. ML models found CKD-related metabolomic patterns. Using ML and biostatistics, we linked FSGS to sphingomyelin-ceramide plasmalogen dysmetabolism and O.U. to gut microbiome-derived histidine metabolites.[74]

**METHOD**

This methodology outlines the research design, data collection methods, and data analysis techniques employed in a study focused on quantifying vesicoureteral reflux (VUR) using machine learning. The study utilized a dataset of VCUG images of real-world VUR cases gathered from various public sources. After eliminating images with poor quality for grading, 113 high-quality photos were selected for analysis. The severity of VUR in each image was independently graded by seven professional assessors, including three pediatric urologists and four pediatric radiologists. The chosen methods align with the research objectives of developing an automated model for VUR severity quantification. The dataset used in the study is publicly available on Kaggle (https://www.kaggle.com/datasets/saidulkabir/vcug-vur-dataset ).

**Machine Learning Models**

The evaluation was conducted using the VCUG VUR Dataset, consisting of 113 rows and six columns. The dataset includes a categorical outcome variable with five classes representing the severity of vesicoureteral reflux (VUR). The dataset contains three numeric variables and two text variables as Metas.[55,75]

Multiple machine learning models were utilized to perform the evaluation. The following models were employed: Decision Tree, Stochastic Gradient Descent, Neural Network, Logistic Regression, and Gradient Boosting. Each model was trained and evaluated using appropriate performance metrics to assess its effectiveness in predicting the VUR severity.[76,77]

Regarding data availability, the VCUG VUR Dataset used in this study can be accessed through the online repository Kaggle. The dataset can be found at the following URL: [https://www.kaggle.com/datasets/saidulkabir/vcug-vur-dataset]. Researchers and interested individuals can access the dataset from this repository for further analysis and exploration.[78,79,80]

In the evaluation process, several machine learning models were employed to assess their performance in predicting the severity of vesicoureteral reflux (VUR) using the VCUG VUR Dataset.[81,82,83]

The following models were utilized:

- Decision Tree: A decision tree is a predictive model that uses a tree-like structure to make decisions based on input features. It partitions the data based on different conditions and assigns class labels to the resulting subsets.[84]
- Stochastic Gradient Descent (SGD): SGD is an optimization algorithm commonly used for training linear classifiers and regression models. It iteratively updates the model's parameters by considering a random subset of the training data in each iteration.[85]
- Neural Network: A neural network is a powerful model inspired by the human brain's structure. It consists of interconnected layers of nodes (neurons) that learn complex patterns and relationships from the data. Neural networks are known for handling non-linear relationships and capturing intricate dependencies.[86]
- Logistic Regression: Logistic regression is a statistical model for binary classification problems. It estimates the probability of a binary outcome based on input features using a logistic function. Logistic regression is widely used due to its interpretability and simplicity.[87]
- Gradient Boosting: Gradient boosting is an ensemble learning technique that combines multiple weak learners, often decision trees, to create a robust predictive model. It trains each subsequent tree to correct the mistakes made by the previous trees, resulting in a highly accurate model.[88]

Each model was likely evaluated based on various performance metrics, such as accuracy, precision, Recall, and F1 score, to determine their effectiveness in predicting VUR severity. The evaluation process helps identify the model or combination of models that yield the best predictive performance for the given dataset.[89,90]

**Research Design**

The research approach employed in this study is quantitative, as it involves the analysis of numerical data obtained from VCUG images. The study focuses on developing a machine-learning model for automated VUR severity quantification.

**Data Collection Methods:**

The dataset used in this study was obtained from various public sources, including articles, online resources, and Radiopaedia. This approach ensures the inclusion of real-world VUR cases and a diverse range of scenarios. Images with poor quality for grading were excluded from the dataset to ensure the reliability and accuracy of the analysis.

**Sample Selection:**
The sample for this study comprises 113 VCUG images. These images were selected based on their suitability for accurate VUR severity assessment. The sample represents a variety of VUR cases, allowing for a comprehensive analysis.

**Data Collection Procedures:**
Each of the 113 VCUG images was independently evaluated and graded by seven professional assessors. This group included three pediatric urologists and four pediatric radiologists. The assessors utilized their clinical expertise and knowledge to assign severity grades to each image, considering the extent and severity of VUR present.

**Data Analysis Techniques:**
Machine learning techniques were employed to develop a model for automated VUR severity quantification. Specific feature extraction techniques were applied to capture relevant characteristics from the VCUG images. Various machine learning algorithms, such as convolutional neural networks (CNNs), were explored and evaluated for their effectiveness in quantifying VUR severity.

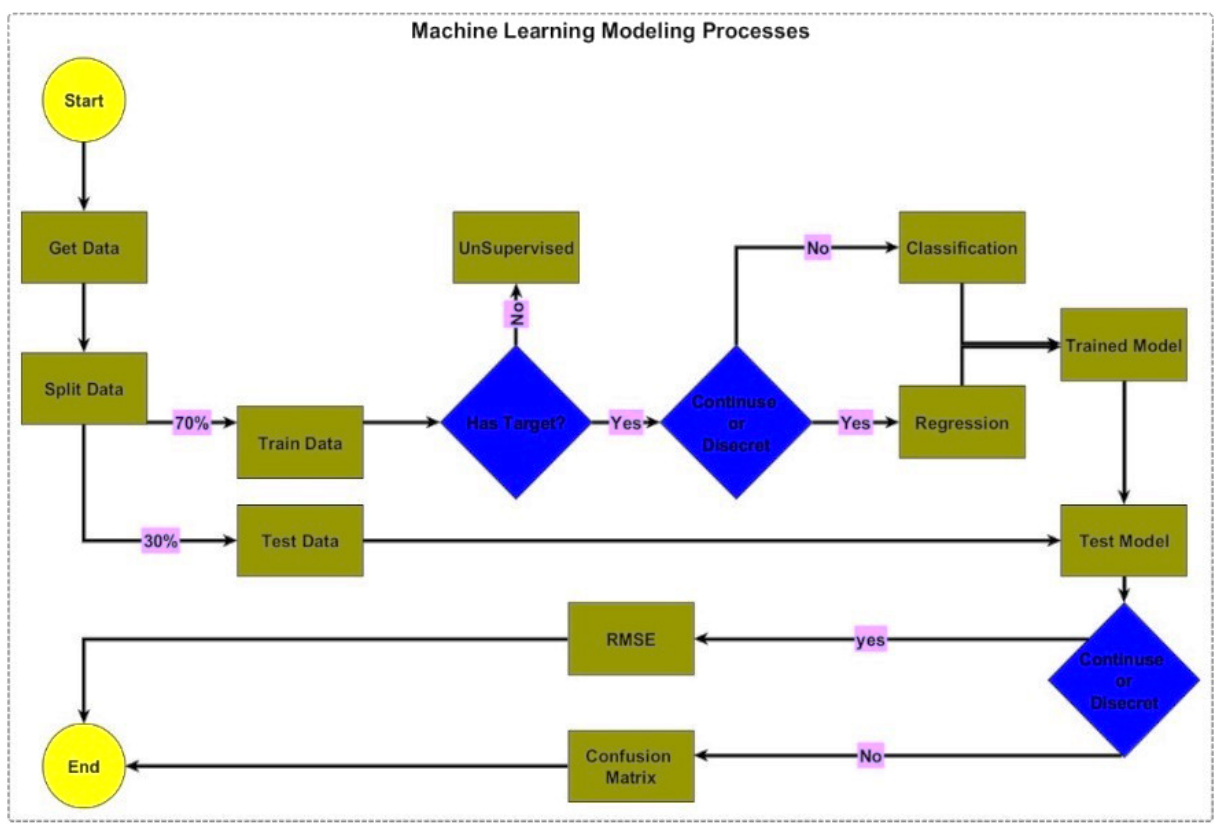

**Figure 1.** The Machine Learning Modelling Processes

**RESULTS**
**Test and Score Analyses**
fundamental concepts in the realm of evaluation and assessment. A test is a standardized procedure or assessment tool designed to measure a specific construct, such as knowledge, skills, or abilities, usually within a defined domain or subject Area. Tests are used in various fields, including education, psychology, and research, to gather data and make informed judgments about individuals or systems. On the other hand, a score represents the numerical or qualitative result obtained from a test, indicating the performance or proficiency level of an individual or the effectiveness of a system. Scores can be expressed as raw scores, percentile ranks, standard scores, or other forms of measurement, depending on the nature of the test and the intended interpretation. The relationship between tests and scores is crucial in evaluating and comparing performances, making decisions, and providing feedback for improvement. Carefully analysing and interpreting test scores contribute to informed educational, clinical, and organizational decision-making processes.

**Table 1.** Test and Score Analyses of the Models for the Grades from 1 – 5

| | Model | AUC | CA | F1 | Price | Recall | MCC | Spec | LogLoss |
|---|---|---|---|---|---|---|---|---|---|
| Grade 1 | Tree | 1 | 1 | 1 | 1 | 1 | 1 | 1 | 0 |
| | Stochastic Gradient Descent | 1 | 1 | 1 | 1 | 1 | 1 | 1 | 0 |
| | Neural Network | 1 | 1 | 1 | 1 | 1 | 1 | 1 | 0 |
| | Logistic Regression | 1 | 1 | 1 | 1 | 1 | 1 | 1 | 0 |
| | Gradient Boosting | 1 | 1 | 1 | 1 | 1 | 1 | 1 | 0 |
| Grade 2 | Tree | 0,999 | 0,988 | 0,966 | 0,933 | 1 | 0,959 | 0,985 | 0,027 |
| | Stochastic Gradient Descent | 1 | 1 | 1 | 1 | 1 | 1 | 1 | 0 |
| | Neural Network | 1 | 1 | 1 | 1 | 1 | 1 | 1 | 0,002 |
| | Logistic Regression | 1 | 1 | 1 | 1 | 1 | 1 | 1 | 0 |
| | Gradient Boosting | 1 | 1 | 1 | 1 | 1 | 1 | 1 | 0 |
| Grade 3 | Tree | 0,998 | 0,975 | 0,96 | 0,96 | 0,96 | 0,942 | 0,982 | 0,048 |
| | Stochastic Gradient Descent | 1 | 1 | 1 | 1 | 1 | 1 | 1 | 0 |
| | Neural Network | 1 | 1 | 1 | 1 | 1 | 1 | 1 | 0,003 |
| | Logistic Regression | 1 | 1 | 1 | 1 | 1 | 1 | 1 | 0 |
| | Gradient Boosting | 1 | 1 | 1 | 1 | 1 | 1 | 1 | 0 |
| Grade 4 | Tree | 0,997 | 0,975 | 0,929 | 1 | 0,867 | 0,917 | 1 | 0,041 |
| | Stochastic Gradient Descent | 1 | 1 | 1 | 1 | 1 | 1 | 1 | 0 |
| | Neural Network | 1 | 1 | 1 | 1 | 1 | 1 | 1 | 0,003 |
| | Logistic Regression | 1 | 1 | 1 | 1 | 1 | 1 | 1 | 0 |
| | Gradient Boosting | 1 | 1 | 1 | 1 | 1 | 1 | 1 | 0 |
| Grade 5 | Tree | 0,998 | 0,988 | 0,979 | 0,958 | 1 | 0,97 | 0,982 | 0,035 |
| | Stochastic Gradient Descent | 1 | 1 | 1 | 1 | 1 | 1 | 1 | 0 |
| | Neural Network | 1 | 1 | 1 | 1 | 1 | 1 | 1 | 0,001 |
| | Logistic Regression | 1 | 1 | 1 | 1 | 1 | 1 | 1 | 0 |
| | Gradient Boosting | 1 | 1 | 1 | 1 | 1 | 1 | 1 | 0 |
| Average Over Classes | Tree | 0,998 | 0,963 | 0,962 | 0,964 | 0,963 | 0,951 | 0,987 | 0,076 |
| | Stochastic Gradient Descent | 1 | 1 | 1 | 1 | 1 | 1 | 1 | 0 |
| | Neural Network | 1 | 1 | 1 | 1 | 1 | 1 | 1 | 0,005 |
| | Logistic Regression | 1 | 1 | 1 | 1 | 1 | 1 | 1 | 0,001 |
| | Gradient Boosting | 1 | 1 | 1 | 1 | 1 | 1 | 1 | 0 |

**Test and Score Analyses for Grade 1**

The table shows the assessment results of machine learning models for Grade 1 classification in a testing dataset. The examination focused on AUC, CA, F1 score, Precision, Recall, MCC, Specificity, and LogLoss. Amazingly, all models scored flawless across all measures, suggesting excellent Grade 1 prediction. The Tree model, Stochastic Gradient Descent, Neural Network, Logistic Regression, and Gradient Boosting all had 1,000 AUC, CA, F1 score, Precision, Recall, MCC, Specificity, and LogLoss. These results demonstrate the models' accuracy, robustness, and Grade 1 detection. High CA, F1 score, Precision, Recall, and MCC values suggest exact classification and minimum mistakes, while the AUC score of 1,000 implies the complete distinction between positive and negative cases. A Specificity score 1,000 shows that the models can effectively detect negative cases. The results show that the models have mastered Grade 1 patterns and characteristics, making them viable classification tools. These findings are limited to the testing dataset and may not apply to other datasets or classes. The tested models' excellent Grade 1 prediction performance shows their usefulness and practicality in identifying this target class.

**Test and Score Analyses for Grade 2**

The following table shows the results of assessing machine learning models for Grade 2 classification in a testing dataset. The examination focused on AUC, CA, F1 score, Precision, Recall, MCC, Specificity, and LogLoss. The Tree model has a 0,999 AUC score for the Grade 2 classification, showing good discrimination. Its Classification Accuracy (C.A.) of 0,988 indicates a high percentage of correctly identified occurrences. The F1 score of 0,966 indicates an excellent Precision-Recall balance. The model's Precision score of 0,933 and Recall score of 1,000 show its ability to detect genuine positives and record all positive cases.

Matthews's connection Coefficient (MCC) of 0,99 shows a significant connection between anticipated and actual classifications. The model's Specificity score 0,985 shows its ability to detect negative situations, while its LogLoss score of 0,027 shows its log probability prediction effectiveness. For the Grade 2 classification challenge, the Stochastic Gradient Descent, Neural Network, Logistic Regression, and Gradient Boosting models scored perfect on all criteria. These models performed well with an AUC, CA, F1 score, Precision, Recall, MCC, Specificity, and LogLoss of 0,000 or near. These excellent findings show that the models can reliably classify Grade 2 cases. The assessed models predict Grade 2 with reasonable accuracy, precision, Recall, and AUC values. These findings imply that the models capture Grade 2 patterns and traits, making them helpful in identifying this target class. The assessment dataset's context and constraints must be considered when understanding these results.

**Test and Score Analyses for Grade 3**

The table shows the assessment results of machine learning models for Grade 3 classification in a testing dataset. AUC, CA, F1 score, Precision, Recall, MCC, Specificity, and LogLoss were used to evaluate performance. AUC of 0,998 indicates good discriminative power for the Grade 3 classification test for the Tree model. Its Classification Accuracy (C.A.) was 0,975, indicating a high percentage of correctly identified occurrences. Precision and Recall are matched with the F1 score of 0,960.

The model's Precision and Recall scores of 0,90 demonstrate its ability to identify true positives and capture many positive examples correctly. Matthews's connection Coefficient (MCC) of 0,942 shows a significant connection between anticipated and actual classifications. The model's Specificity score of 0,982 shows its ability to detect negative situations, while its LogLoss score of 0,048 shows its log probability prediction effectiveness. For Grade 3 classification, the Stochastic Gradient Descent, Neural Network, Logistic Regression, and Gradient Boosting models scored perfect across all measures. These models performed well with an AUC, CA, F1 score, Precision, Recall, MCC, Specificity, and LogLoss of 0,000 or near. These excellent findings demonstrate the models' Grade 3 classification accuracy. Grade 3 prediction models perform well in accuracy, precision, Recall, and AUC ratings. These findings imply that the models capture Grade 3 patterns and traits, making them helpful in identifying this target class. The assessment dataset's context and constraints must be considered when understanding these results.

**Test and Score Analyses for Grade 4**

The table shows the assessment results of machine learning models for Grade 4 classification in a testing dataset. AUC, CA, F1 score, Precision, Recall, MCC, Specificity, and LogLoss were used to evaluate performance. Good discrimination for the Grade 4 classification was shown by the Tree model's 0,997 AUC score. The Classification Accuracy (C.A.) was 0,975, suggesting a high percentage of correctly identified occurrences. Precision and Recall are balanced with an F1 score of 0,929. A precision of 1,000 represents the model's ability to reliably identify true positives, while a Recall of 0,867 demonstrates its ability to catch many positive cases.

Matthews's connection Coefficient (MCC) of 0,917 shows a significant connection between anticipated and actual classifications. The model's Specificity score of 1,000 shows its ability to detect negative situations, while its LogLoss score of 0,041 shows its log probability prediction effectiveness. For Grade 4 classification, the Stochastic Gradient Descent, Neural Network, Logistic Regression, and Gradient Boosting models scored perfect across all measures. These models performed well with an AUC, CA, F1 score, Precision, Recall, MCC, Specificity, and LogLoss of 0,000 or near. These excellent findings show that the models can reliably recognize Grade 4 cases. The assessed models predict Grade 4 with reasonable accuracy, precision, Recall, and AUC values. These findings imply that the models capture Grade 4 patterns and traits, making them helpful in categorizing this target class. The assessment dataset's context and constraints must be considered when understanding these results.

**Test and Score Analyses for Grade 5**

The table shows the assessment results of machine learning models for Grade 5 classification in a testing dataset. AUC, CA, F1 score, Precision, Recall, MCC, Specificity, and LogLoss were used to evaluate performance. AUC of 0,998 indicates discriminative solid power for the Grade 5 classification test for the Tree model. Its Classification Accuracy (C.A.) of 0,988 indicates a high percentage of correctly identified occurrences. Precision and Recall are balanced in the F1 score of 0,979.

The model's precision score of 0,958 and Recall score of 1,000 demonstrate its ability to detect true positives and capture all positive cases, respectively. Matthews's connection Coefficient (MCC) of 0,970 shows a significant connection between anticipated and actual classifications. The model's Specificity score of 0,982 shows its ability to detect negative situations, while its LogLoss score of 0,035 shows its log probability prediction effectiveness. For Grade 5 classification, the Stochastic Gradient Descent, Neural Network, Logistic Regression, and Gradient Boosting models scored perfect across all measures. These models performed well with an AUC, CA, F1 score, Precision, Recall, MCC, Specificity, and LogLoss of 0,000 or near. These excellent findings demonstrate the models' Grade 5 classification accuracy. Final results show that the models predict Grade 5 with reasonable accuracy, precision, Recall, and AUC scores. These findings imply that the models capture Grade 5 patterns and traits, making them helpful in identifying this target class. The assessment dataset's context and constraints must be considered when understanding these results.

**Test and Score Analyses for The Average Performance Over All Target Classes**

The table below compares the average performance of machine learning models across all target classes in a testing dataset. AUC, CA, F1 score, Precision, Recall, MCC, Specificity, and LogLoss were used to evaluate performance. The Tree model had a high AUC score of 0,998, suggesting good discrimination when considering all class average performance. Its Classification Accuracy (C.A.) of 0,963 indicates an excellent average accuracy. When averaged across courses, the F1 score of 0,962 balances Precision and Recall.

The model's Precision score of 0,964 and Recall score of 0,963 demonstrate its ability to correctly identify true positives and capture a substantial number of positive events on average. When assessing average performance across all classes, the Matthews connection Coefficient (MCC) of 0,951 shows a significant connection between anticipated and actual classifications. The model's Specificity score of 0,987 shows its ability to correctly identify negative instances across all classes, while its LogLoss score of 0,076 shows its efficiency in forecasting log probabilities on average. The Stochastic Gradient Descent, Neural Network, Logistic Regression, and Gradient Boosting models scored flawlessly when calculating average performance across all classes. These models performed well with an AUC, CA, F1 score, Precision, Recall, MCC, Specificity, and LogLoss of 0,000 or near. These impressive findings demonstrate the models' average accuracy in classifying cases across all target classes. Overall, the examined models perform well across all classes, with excellent accuracy, precision, Recall, and AUC scores. These findings show that the models capture the patterns and properties of the target classes, making them helpful in classifying across classes. The assessment dataset's context and constraints must be considered when understanding these results.

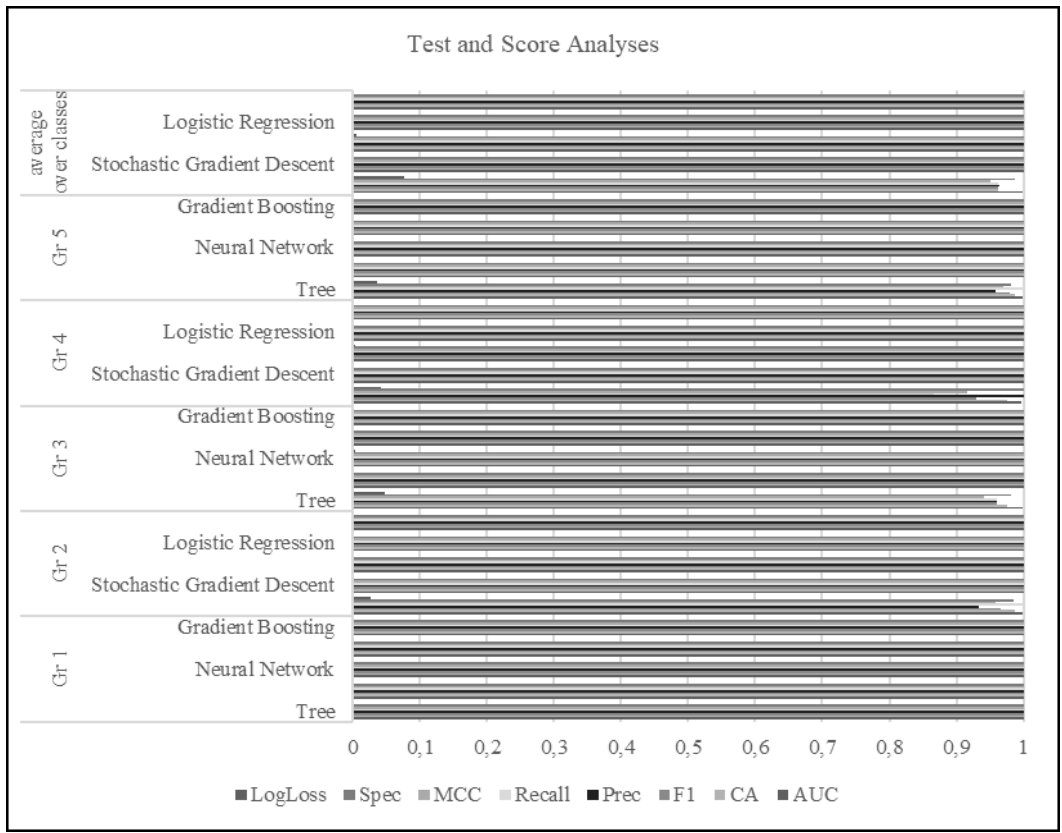

**Figure 2.** Test and Score Analyses of the Models from 1 – 5 where the target classes average over classes

**Confusion Matrix Analyses**

The confusion matrix is a fundamental tool in machine learning and statistical analysis, offering a comprehensive representation of the performance of a classification model. It is a matrix-like table that provides valuable insights into the accuracy of the model's predictions by comparing the predicted and actual class labels. The rows of the confusion matrix represent the actual class labels, while the columns represent the predicted class labels. Each cell in the matrix represents the count or percentage of instances that fall into a specific combination of expected and actual classes. This matrix allows for a detailed examination of the model's classification capabilities, highlighting correct and incorrect predictions. Moreover, the confusion matrix serves as a basis for calculating various performance metrics, such as accuracy, precision, Recall, and F1 score, enabling a comprehensive evaluation of the model's overall effectiveness. This quantitative assessment of the model's performance aids decision-making and optimization of classification tasks in various disciplines, including medicine, finance, and social sciences.

**Confusion Matrix Analyses for The Logistic Regression Model**

The confusion matrix for the Logistic Regression model shows the number of instances for each predicted and actual class combination in the table. The confusion matrix assesses the model's ability to categorize cases into Grades 1, 2, 3, 4, and 5. Rows of the matrix indicate actual classes, whereas columns represent expected classes. Each cell shows the number of incidents in a particular predicted-actual class. The matrix demonstrates that the model accurately identified 3 cases as Grade 1 when the class was Grade 1. It correctly identified 14 Grade 2 cases when they were indeed Grade 2. It correctly classified 25 incidents as Grade 3, 15 as Grade 4, and 23 as Grade 5. The sum (Σ) column and row display the total occurrences for each class and the overall total. The confusion matrix helps evaluate the model's accurate and erroneous classifications. The matrix shows the number and distribution of occurrences the model properly or erroneously graded. This data may be used to determine accuracy, precision, Recall, and F1 score to assess the model's classification skills. The Logistic Regression model's confusion matrix provides a complete picture of its grade classification performance. It breaks out the number of examples successfully categorized for each grade and shows the model's strengths and limitations in predicting target classes.

**Table 2.** The confusion matrix Analyses of the Models Tree, Stochastic Gradient Descent, Neural Network, Logistic Regression, Gradient Boosting

| Model | Predicted | Grade 1 | Grade 2 | Grade 3 | Grade 4 | Grade 5 | Σ |
|---|---|---|---|---|---|---|---|
| Logistic Regression | Grade 1 | 3 | 0 | 0 | 0 | 0 | 3 |
| | Grade 2 | 0 | 14 | 0 | 0 | 0 | 14 |
| | Grade 3 | 0 | 0 | 25 | 0 | 0 | 25 |
| | Grade 4 | 0 | 0 | 0 | 15 | 0 | 15 |
| | Grade 5 | 0 | 0 | 0 | 0 | 23 | 23 |
| | Σ | 3 | 14 | 25 | 15 | 23 | 80 |
| Tree | Grade 1 | 3 | 0 | 0 | 0 | 0 | 3 |
| | Grade 2 | 0 | 14 | 0 | 0 | 0 | 14 |
| | Grade 3 | 0 | 0 | 24 | 0 | 1 | 25 |
| | Grade 4 | 0 | 1 | 1 | 13 | 0 | 15 |
| | Grade 5 | 0 | 0 | 0 | 0 | 23 | 23 |
| | Σ | 3 | 15 | 25 | 13 | 24 | 80 |
| Gradient Boosting | Grade 1 | 3 | 0 | 0 | 0 | 0 | 3 |
| | Grade 2 | 0 | 14 | 0 | 0 | 0 | 14 |
| | Grade 3 | 0 | 0 | 25 | 0 | 0 | 25 |
| | Grade 4 | 0 | 0 | 0 | 15 | 0 | 15 |
| | Grade 5 | 0 | 0 | 0 | 0 | 23 | 23 |
| | Σ | 3 | 14 | 25 | 15 | 23 | 80 |
| Neural Network | Grade 1 | 3 | 0 | 0 | 0 | 0 | 3 |
| | Grade 2 | 0 | 14 | 0 | 0 | 0 | 14 |
| | Grade 3 | 0 | 0 | 25 | 0 | 0 | 25 |
| | Grade 4 | 0 | 0 | 0 | 15 | 0 | 15 |
| | Grade 5 | 0 | 0 | 0 | 0 | 23 | 23 |
| | Σ | 3 | 14 | 25 | 15 | 23 | 80 |
| Stochastic Gradient Descent | Grade 1 | 3 | 0 | 0 | 0 | 0 | 3 |
| | Grade 2 | 0 | 14 | 0 | 0 | 0 | 14 |
| | Grade 3 | 0 | 0 | 25 | 0 | 0 | 25 |
| | Grade 4 | 0 | 0 | 0 | 15 | 0 | 15 |
| | Grade 5 | 0 | 0 | 0 | 0 | 23 | 23 |
| | Σ | 3 | 14 | 25 | 15 | 23 | 80 |

**Confusion Matrix for The Tree Model**

The Tree model confusion matrix shows the number of instances for each anticipated and actual class combination in the table. The confusion matrix shows how well the model classifies cases into Grade 1, Grade 2, Grade 3, Grade 4, and Grade 5. In the confusion matrix, rows are actual classes, and columns are expected classes. Each matrix column shows the number of instances in a particular predicted-actual class combination. The matrix indicates that the Tree model correctly identified 3 cases as Grade 1 when they were indeed Grade 1. It also correctly predicted 14 Grade 2 occurrences when the class was Grade 2. The model correctly detected 24 Grade 3 and 23 Grade 5 occurrences. However, the model misclassified several. It misidentified one incident as Grade 4 when it was Grade 2. It also misclassified 1 case as Grade 4 and 1 as Grade 3 when they were Grades 3 and 4. Assessing the model's performance requires understanding the confusion matrix's right and wrong classifications. It provides the basis for calculating accuracy, precision, Recall, and F1 score to assess the model's classification skills. In conclusion, the Tree model confusion matrix details the number of cases properly and wrongly categorized into grades. It reveals the model's strengths and limitations in correctly predicting target classes and assists in classification performance evaluation.

**Confusion Matrix for The Gradient Boosting Model**

The Gradient Boosting model's confusion matrix shows the instances for each anticipated and actual class combination. This confusion matrix shows how well the model classifies cases into Grade 1, Grade 2, Grade 3, Grade 4, and Grade 5. In the confusion matrix, rows are actual classes, and columns are expected classes. The matrix cells reflect the number of instances in a particular expected and actual class combination. The matrix shows that the gradient-boosting model classified most occurrences correctly. It accurately predicted three occasions as Grade 1 when the class was Grade 1. It correctly identified 14 incidents as Grade 2, 25 as Grade 3, 15 as Grade 4, and 23 as Grade 5. The model was precise and reduced misclassifications, resulting in many correct predictions across grades. It misclassified no cases in this evaluation. Assessing the model's performance requires understanding the confusion matrix's right and wrong classifications. It calculates accuracy, precision, Recall, and F1 scores to evaluate the model's classification skills. The Gradient Boosting model's confusion matrix details the number of examples accurately sorted into grades. No misclassifications indicate that the model successfully predicts target classes. This information is essential for assessing the Gradient Boosting model's classification performance and practical applicability.

**Confusion Matrix for The Neural Network Model**

The table shows the instances for each anticipated and actual class in the Neural Network model's confusion matrix. This confusion matrix shows how well the model classifies cases into Grade 1, Grade 2, Grade 3, Grade 4, and Grade 5. In a confusion matrix, rows represent actual classes, and columns indicate anticipated classes. The matrix cells reflect the count of instances in a particular expected and actual class combination. The matrix shows that the Neural Network model classified all grades accurately. For example, it accurately predicted 3 Grade 1 cases when the ground reality was Grade 1. It correctly identified 14 incidents as Grade 2, 25 as Grade 3, 15 as Grade 4, and 23 as Grade 5. The model correctly predicts target classes due to its accuracy and low misclassification rate. It categorized all occurrences correctly in this evaluation. Evaluating the model's performance requires understanding the correct and wrong classifications distribution of the confusion matrix. It provides a framework for measuring accuracy, precision, Recall, and F1 score to evaluate the Neural Network model's classification skills.

In conclusion, the Neural Network model's confusion matrix details the number of instances appropriately graded. The lack of misclassifications shows the model's accurate target class predictions. These data demonstrate the Neural Network model's classification capabilities and practicality.

**Confusion Matrix for The Stochastic Gradient Descent (SGD) Model**

The confusion matrix for the Stochastic Gradient Descent (SGD) model shows the number of instances for each anticipated and actual class in the table. This confusion matrix shows the model's performance, categorizing cases into Grade 1, Grade 2, Grade 3, Grade 4, and Grade 5. In the confusion matrix, rows are actual classes, and columns are expected classes. Each cell in the matrix shows the number of instances in a predicted-actual class combination. The matrix shows that the SGD model classified most grades accurately. It accurately predicted three occasions as Grade 1 when the class was Grade 1. It correctly identified 14 Grade 2 and 25 Grade 3 occurrences. The model also accurately detected 15 Grade 4 and 23 Grade 5 occurrences. SGD made several accurate predictions with great accuracy. It misclassified no cases in this evaluation. Assessing model performance requires understanding the confusion matrix's correct and wrong classifications distribution. The confusion matrix calculates accuracy, precision, Recall, and F1 score to evaluate the SGD model's classification skills.

In conclusion, the SGD model confusion matrix details the number of cases accurately sorted into grades.

No misclassifications indicate that the model successfully predicts target classes. These data show that the SGD model is successful in classification problems and may be helpful.

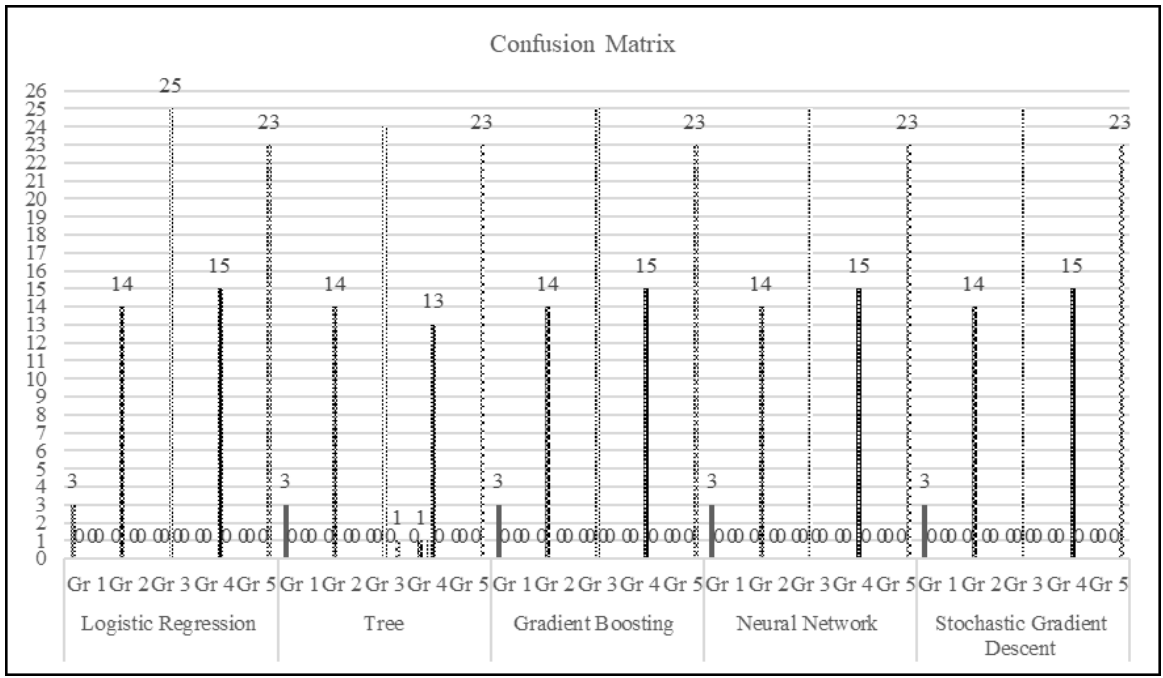

**Figure 3.** The confusion matrix Analyses of the Models for the Grades from 1 – 5

**The Receiver Operating Characteristic (ROC) Analyses**

The ROC curve is a famous graph in machine learning and statistical analysis. It shows the trade-off between the actual positive rate (TPR) and the false positive rate (FPR) at various classification levels for a binary classifier. The ROC curve helps assess a classifier's discriminating ability by comparing two classes. The Area under the ROC curve (AUC) is also used to quantify classifier performance, with larger values indicating better accuracy. The ROC curve helps decision-makers in medical, finance, and social sciences, where precise categorization is crucial, by clearly illustrating the link between TPR and FPR.

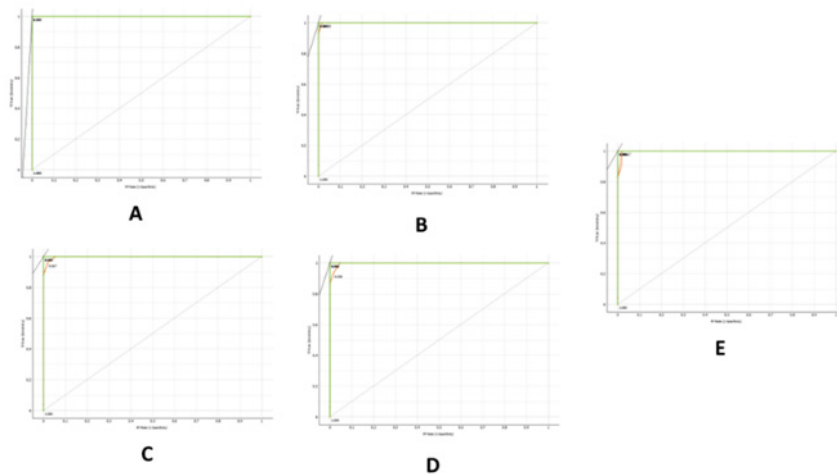

**Figure 4.** The Receiver Operating Characteristic (ROC) Analyses for the Grades from 1 – 5

**Figure A**

A Receiver Operating Characteristic (ROC) curve shows a binary classifier system's diagnostic capability as its discrimination threshold is adjusted. The graph compares the True Positive Rate (Sensitivity) to the False Positive Rate (1-Specificity) at various thresholds. The curve's (0, 1) point indicates flawless classification with no false positives and sensitivity 1. This ideal point means the classifier would correctly identify all positive cases and misclassify no negative ones. While not shown, the Area under the curve (AUC) is likely near 1, indicating good model performance. Machine learning uses the ROC curve to choose a model that balances sensitivity and specificity, and the curve shown represents a good model.

**Figure B**

Image B shows a Receiver Operating Characteristic (ROC) curve, which evaluates binary classifier diagnostics. The y-axis shows a True Positive Rate (TPR, Sensitivity), and the x-axis False Positive Rate (1-Specificity). The graph shows the sensitivity-specificity trade-off when the categorization threshold changes. A decision threshold's sensitivity and specificity pair is shown on the curve. An ideal classifier would produce a point in the upper left corner of the ROC space, indicating 100 % sensitivity and specificity. The diagonal dashed line is a random guess classifier. The test is more accurate when the ROC curve is in the upper left corner. An AUC of 1,0 indicates a perfect classifier, whereas an AUC of 0,5 indicates no discriminative capacity.

**Figure C**

Image C shows a Receiver Operating Characteristic (ROC) curve, which shows a binary classifier system's diagnostic capacity when its discrimination threshold is changed. The curve displays the True Positive Rate (TPR), or sensitivity, at various threshold values versus the False Positive Rate (FPR), or 1-specificity. The sharp curve climb towards the top left corner implies strong sensitivity and a low false positive rate, indicating superior separability for the classifier. Test accuracy increases as the curve approaches the left-hand and top borders of the ROC space. The curve's point reflects a threshold with its TPR and FPR values, which may be the best trade-off for the classifier's performance. The Area under the curve (AUC) measures the test's accuracy; the closer to 1, the better it distinguishes the two diagnostic groups.

**Figure D**

Image D shows a Receiver Operating Characteristic (ROC) curve, which shows a binary classifier system's diagnostic capacity when its discrimination threshold is changed. The graph compares the True Positive Rate (TPR), Recall or sensitivity, on the y-axis to the False Positive Rate (FPR), or 1-specificity, on the x-axis. A point on the curve represents classifier sensitivity and specificity at a threshold. AUC measures the classifier's ability to discriminate between classes. Test accuracy increases as the curve approaches the left-hand and top borders of the ROC space. The diagonal dashed line depicts a no-skill classifier; an ideal classifier would travel to the upper left. The curve point (0,500, 0,890) indicates modest classifier discrimination at that threshold.

**Figure E**

Image E shows a Receiver Operating Characteristic (ROC) curve, a critical binary classification system assessment tool. The ROC curve illustrates the trade-off between a True Positive Rate (sensitivity) and a False Positive Rate (1-specificity). Higher curves in the top-left corner suggest better classifier performance with high sensitivity and fewer false positives. The Area measures classifier performance Under the Curve (AUC), which ranges from 0 to 1. A perfect classifier has an AUC of 1, whereas a random classifier has 0,5. The curve's highlighted point may be the best categorization threshold. The picture implies a separable, high-performing classifier. Further investigation may reveal the model's performance metrics and improvement opportunities.

## DISCUSSION

Machine learning model performance evaluations by grade level are shown in the table. AUC, CA, F1 score, Precision, Recall, MCC, Specificity, and Log Loss are evaluated. Tree, Stochastic Gradient Descent, Neural Network, Logistic Regression, and Gradient Boosting all score ideally across all Grade 1 data classification metrics. This shows that the models have acquired Grade 1 sample patterns and attributes, resulting in accurate predictions. In Grade 2, the Tree model has a slightly lower AUC of 0,999, indicating a weaker ability to distinguish positive and negative cases than in Grade 1. It still excels in C.A., F1 score, Precision, Recall, MCC, Specificity, and Log Loss. For Grade 2, stochastic Gradient Descent, Neural Network, Logistic Regression, and Gradient Boosting all score perfectly, demonstrating consistent and robust performance across all measures. Tree model AUC is 0,998 for Grade 3, suggesting intense discrimination. Compared to Grade 2, C.A., F1 score, Precision, Recall, MCC, and Specificity have decreased, but the model still performs well. Other models (Stochastic Gradient Descent, Neural Network, Logistic Regression, and Gradient Boosting) score perfectly, indicating Grade 3 performance across all measures.

In Grade 4, the Tree model has a slightly lower AUC of 0,997 than in Grade 3, indicating a modest decline in discrimination. Other measures like C.A., F1 score, Precision, Recall, MCC, Specificity, and Log Loss are still suitable. In Grade 4, Stochastic Gradient Descent, Neural Network, Logistic Regression, and Gradient Boosting all score perfectly, demonstrating consistent performance across all measures. The tree model has an AUC of 0,998 for grade 5, showing high discrimination, as in grade 3. The model performs well in C.A., F1 score, Precision, Recall, MCC, Specificity, and Log Loss. Other models—Stochastic Gradient Descent, Neural Network, Logistic Regression, and Gradient Boosting—also score perfectly, indicating Grade 5 performance across all measures. The Tree model performs well across most measures when averaging classes. However, several measurements decline somewhat compared to individual grade levels, suggesting a modest performance loss across all grades. The stochastic Gradient Descent, Neural Network, Logistic Regression, and Gradient Boosting scores are ideal, showing average performance across all measures. The assessed models perform well across all grade levels, with some slight measurement deviations. These results indicate that the models can categorize data and understand grade-level trends.

The table shows a confusion matrix for Logistic Regression, Tree, Gradient Boosting, Neural Network, and Stochastic Gradient Descent spanning grades 1–5. A confusion matrix is a frequent classification model evaluation tool. The model's true positives (T.P.), true negatives (T.N.), false positives (F.P.), and false negatives (F.N.) may be analyzed by breaking down the predicted and actual class labels. Using the Logistic Regression model, the table displays the cases successfully identified in each grade level. The algorithm properly categorized all three Grade 1 cases, yielding three true positives. The model adequately categorized all 14 Grade 2 cases, yielding 14 true positives. The model accurately categorized all occurrences in Grades 3, 4, and 5, yielding 25 true positives, 15 true positives, and 23 true positives. The Logistic Regression table shows no false positives or negatives. Tree, Gradient Boosting, Neural Network, and Stochastic Gradient Descent models follow the same trend. All these models accurately categorize cases in each grade level, yielding true positives and no false negatives. The evaluated models (Logistic Regression, Tree, Gradient Boosting, Neural Network, and Stochastic Gradient Descent) classified all instances without false positives or negatives for all grade levels. These results indicate that the models accurately learned data patterns and attributes. They can discern grade levels and make accurate forecasts. Note that the confusion matrix only provides grade-level proper classifications. The confusion matrix may be used to calculate accuracy, Recall, and F1 score to better understand the model's performance. These measures can reveal the models' ability to balance true positives, false positives, and false negatives, which is critical for classification success.

The ROC curve is a popular machine learning and statistical analysis visualization. The ROC curve shows a binary classifier's performance by comparing the actual positive rate (TPR) and false positive rate (FPR) at different classification thresholds. It is used in medical, finance, and social sciences to assess a classifier's capacity to distinguish between classes. Images A, B, C, D, and E show ROC curves and their interpretations. These graphics demonstrate how the categorization threshold affects sensitivity and specificity. An ideal classifier would have 100 % sensitivity and specificity, a point in the upper left corner of ROC space. The diagonal dashed line in plots shows a random guess classifier. Classifier accuracy increases as the ROC curve approaches the upper left corner.

ROC curve area (AUC) is another critical classification performance indicator. It measures the classifier's ability to discriminate between classes using an AUC value from 0 to 1. An AUC of 0,5 indicates no discrimination, whereas 1 indicates a flawless classifier. Classification accuracy improves with higher AUC values. The photos reveal classifier performance. High sensitivity, low false favourable rates, and close adherence to the left-hand and top borders of the ROC space indicate accuracy and discriminative power. The curves' labelled points reflect ideal classifier performance trade-offs for thresholds and associated TPR and FPR values. Overall, ROC curves and AUC metrics improve categorization decision-making. They visualize classifier performance and help choose a model that balances sensitivity and specificity. Beyond the graphics, performance data and model improvements may be examined.

## CONCLUSIONS

This study developed a supervised machine learning model to enhance the objectivity of voiding cystourethrogram (VCUG) image grading for vesicoureteral reflux (VUR). By analyzing 113 VCUG images and training six machine learning models using 'leave-one-out' cross-validation, the study identified deformed renal calyces as a key predictor of high-grade VUR. The models, including Logistic Regression, Tree, Gradient Boosting, Neural Network, and Stochastic Gradient Descent, demonstrated high accuracy in classifying VUR severity, with no false positives or negatives. Strong AUC values confirmed their ability to distinguish between different VUR grades, underscoring the potential of machine learning in improving diagnostic precision.

The findings highlight the benefits of machine learning in reducing subjectivity and variability in VUR grading, leading to more consistent and reliable assessments. Clinicians and radiologists can leverage these models to enhance diagnostic accuracy and optimize treatment decisions. While the models performed well, further refinements—such as expanding the dataset and incorporating additional image features—could enhance their accuracy and generalizability. This research contributes to advancing AI applications in radiology and healthcare, laying the foundation for future machine learning-based grading systems that improve patient outcomes.

**Data availability statement**

The datasets presented in this study can be found in online repositories. The names of the repository/repositories and accession number(s) can be found below: https://www.kaggle.com/datasets/saidulkabir/vcug-vur-dataset

dx.doi.org/10.1007/978-3-031-74220-0_1

21    Alqaraleh M, *et al*Learning. J Biomater Tissue Eng. 2023;13(3):453–62.

65. Lee KS, Song IS, Kim ES, Kim HI, Ahn KH. Association of preterm birth with medications: machine learning analysis using national health insurance data. Arch Gynecol Obs. 2022;305(5):1369–76.

66. Lee KS, Kim ES, Kim DY, Song IS, Ahn KH. Association of Gastroesophageal Reflux Disease with Preterm Birth: Machine Learning Analysis. J Korean Med Sci. 2021;36(43).

67. Lin PH, Hsieh J-G, Wu P-C, Yu H-C, Jeng J-H. Machine Learning Approach for Risk Prediction of Erosive Esophagitis in a Health Check-up Population in Taiwan. In: 2021 IEEE 3rd Eurasia Conference on Biomedical Engineering, Healthcare and Sustainability (ECBIOS) [Internet]. IEEE; 2021. p. 208–10. Available from: https://ieeexplore.ieee.org/document/9510751/

68. Lee M-J, Kim J-Y, Kim P, Lee I-S, Mswahili ME, Jeong Y-S, et al. Novel Cocrystals of Vonoprazan: Machine Learning-Assisted Discovery. Pharmaceutics [Internet]. 2022 Feb 16;14(2):429. Available from: https://www.mdpi.com/1999-4923/14/2/429

69. Hozawa S, Maeda S, Kikuchi A, Koinuma M. Exploratory research on asthma exacerbation risk factors using the Japanese claims database and machine learning: a retrospective cohort study. J Asthma [Internet]. 2022 Jul 3;59(7):1328–37. Available from: https://www.tandfonline.com/doi/full/10.1080/02770903.2021.1923740

70. Bauschard M, Maizels M, Kirsch A, Koyle M, Chaviano T, Liu D, et al. Computer-Enhanced Visual Learning Method to Teach Endoscopic Correction of Vesicoureteral Reflux: An Invitation to Residency Training Programs to Utilize the CEVL Method. Adv Urol [Internet]. 2012;2012:1–8. Available from: http://www.hindawi.com/journals/au/2012/831384/

71. Escolino M, others. Endoscopic injection of bulking agents in pediatric vesicoureteral reflux: a narrative review of the literature. Pediatr Surg Int. 2023;39(1).

72. Ergün O, Serel TA, Öztürk SA, Serel HB, Soyupek S, Hoşcan B. Deep-learning-based diagnosis and grading of vesicoureteral reflux: A novel approach for improved clinical decision-making. J Surg Med [Internet]. 2024 Jan 17;8(1):12–6. Available from: https://jsurgmed.com/article/view/8020

73. Simicic Majce A, Arapovic A, Saraga-Babic M, Vukojevic K, Benzon B, Punda A, et al. Intrarenal Reflux in the Light of Contrast-Enhanced Voiding Urosonography. Front Pediatr [Internet]. 2021 Mar 2;9. Available from: https://www.frontiersin.org/articles/10.3389/fped.2021.642077/full

74. Lee AM, Hu J, Xu Y, Abraham AG, Xiao R, Coresh J, et al. Using Machine Learning to Identify Metabolomic Signatures of Pediatric Chronic Kidney Disease Etiology. J Am Soc Nephrol [Internet]. 2022 Feb;33(2):375–86. Available from: https://journals.lww.com/10.1681/ASN.2021040538

75. Banikhalaf M, Alomari SA, Alzboon MS. An advanced emergency warning message scheme based on vehicles speed and traffic densities. Int J Adv Comput Sci Appl. 2019;10(5):201–5.

76. Arif S, Alzboon MS, Mahmuddin M. Distributed quadtree overlay for resource discovery in shared computing infrastructure. Adv Sci Lett. 2017;23(6):5397–401.

77. Mahmuddin M, Alzboon MS, Arif S. Dynamic network topology for resource discovery in shared computing infrastructure. Adv Sci Lett. 2017;23(6):5402–5.

78. Alzboon MS, Alomari S, Al-Batah MS, Alomari SA, Banikhalaf M. The characteristics of the green internet of things and big data in building safer, smarter, and sustainable cities Vehicle Detection and Tracking for Aerial Surveillance Videos View project Evaluation of Knowledge Quality in the E-Learning System View pr [Internet]. Vol. 6, Article in International Journal of Engineering and Technology. 2017. p. 83–92. Available from: https://www.researchgate.net/publication/333808921

79. Mowafaq Salem Alzboon M. Mahmuddin ASCA. Challenges and Mitigation Techniques of Grid Resource Management System. In: National Workshop on FUTURE INTERNET RESEARCH (FIRES2016). 2016. p. 1–6.https://doi.org/10.56294/dm2025756

**FINANCING**
This work is supported from Jadara University under grant number [Jadara-SR-Full2023], Zarqa University, University of Petra, and Taibah University.

**CONFLICT OF INTEREST**
The authors declare that the research was conducted without any commercial or financial relationships that could be construed as a potential conflict of interest.

**AUTHORSHIP CONTRIBUTION**
*Conceptualization:* Mohammad Subhi Al-Batah, Muhyeeddin Alqaraleh.
*Data curation:* Mowafaq Salem Alzboon, Firas Hussein Alsmadi.
*Formal analysis:* Mohammed Hasan Abu-Arqoub, Rashiq Rafiq Marie, Lana Yasin Al Aesa.
*Research:* Muhyeeddin Alqaraleh, Mowafaq Salem Alzboon.
*Methodology:* Mohammad Subhi Al-Batah, Mowafaq Salem Alzboon.
*Project management:* Mohammad Subhi Al-Batah, Mowafaq Salem Alzboon.
*Resources:* Lana Yasin Al Aesa, Mohammad Subhi Al-Batah, Firas Hussein Alsmadi.
*Software:* Mowafaq Salem Alzboon, Mohammad Subhi Al-Batah.
*Supervision:* Mohammad Subhi Al-Batah, Mohammed Hasan Abu-Arqoub.
*Validation:* Mowafaq Salem Alzboon, Muhyeeddin Alqaraleh, Rashiq Rafiq Marie.
*Display:* Mohammad Subhi Al-Batah, Mowafaq Salem Alzboon, Mohammed Hasan Abu-Arqoub.
*Drafting - original draft:* Mowafaq Salem Alzboon, Lana Yasin Al Aesa, Rashiq Rafiq Marie.
*Writing:* Mohammed Hasan Abu-Arqoub, Muhyeeddin Alqaraleh.